\documentclass{article} 
\usepackage{iclr2026_conference,times}

\usepackage{amsmath,amsfonts,bm}

\def\eqref#1{equation~\ref{#1}}

\def\1{\bm{1}}

\DeclareMathAlphabet{\mathsfit}{\encodingdefault}{\sfdefault}{m}{sl}
\SetMathAlphabet{\mathsfit}{bold}{\encodingdefault}{\sfdefault}{bx}{n}

\usepackage{hyperref}
\usepackage{subcaption}
\usepackage{url}
\usepackage{booktabs}
\usepackage{graphicx}
\usepackage{makecell}

\usepackage{enumitem} 

\AtBeginDocument{%
  \setlength\abovedisplayskip{4pt}
  \setlength\belowdisplayskip{4pt}
  \setlength\abovedisplayshortskip{2pt}
  \setlength\belowdisplayshortskip{2pt}
}

\title{EO-VAE: Towards A Multi-sensor Tokenizer for Earth Observation Data}

\author{Nils Lehmann, Yi Wang, Zhitong Xiong, \& Xiaoxiang Zhu \\
Data Science in Earth Observation \\
Technical University of Munich (TUM)\\
Munich Center for Machine Learning (MCML) \\
Munich, Germany \\
\texttt{\{n.lehmann,xiaoxiang.zhu\}@tum.de} \\
}

\iclrfinalcopy 
\begin{document}

\maketitle


\begin{abstract}
State-of-the-art generative image and video models rely heavily on tokenizers that compress high-dimensional inputs into more efficient latent representations. While this paradigm has revolutionized RGB generation, Earth observation (EO) data presents unique challenges due to diverse sensor specifications and variable spectral channels. We propose EO-VAE, a multi-sensor variational autoencoder designed to serve as a foundational tokenizer for the EO domain. Unlike prior approaches that train separate tokenizers for each modality, EO-VAE utilizes a single model to encode and reconstruct flexible channel combinations via dynamic hypernetworks. Our experiments on the TerraMesh dataset demonstrate that EO-VAE achieves superior reconstruction fidelity compared to the TerraMind tokenizers, establishing a robust baseline for latent generative modeling in remote sensing.
\end{abstract}

\section{Introduction}
The Stable Diffusion generative model introduced by \citet{rombach2022high} was a central breakthrough in high resolution image generation. Both training and inference efficiency, as well as overall generation performance rely on a pretrained Variational Autoencoder \citep{VAE} that has since become a fundamental building block and idea in numerous domains like video generation \citep{brooks2024video}, weather forecasting \citep{nguyen2025omnicast}, and newer text-to-image models \citep{flux2}.

In contrast to RGB data, processing earth observation data poses several challenges such as non-fixed pixel value ranges, multispectral channels, sensor diversity and large data volumes. With the growing volume of available earth observation data, reaching the petabyte scale, latent modeling approaches have similarly appealing properties of reducing memory requirements and improving efficiency. Previous works of earth observation generative models like \citet{khanna2024diffusionsat} and \citet{jakubik2025terramind} also make use of pretrained Autoencoders but have limitations. \cite{khanna2024diffusionsat} use the pretrained SD-VAE model which works extremely well in the RGB domain but cannot operate on channel varying satellite imagery. In contrast, \citet{jakubik2025terramind} train a separate tokenizer based on the ViT-VQGAN \citep{yuvector} framework for each of the separate modalities contained in the Terramesh dataset \citep{blumenstiel2025terramesh}. We instead propose EO-VAE, a single autoencoder model that can encode and reconstruct a variable number of channels conditioned on the channel wavelengths. Our experiments demonstrate superior reconstruction performance compared to the TerraMind tokenizers. \footnote{Code is available at \href{https://github.com/nilsleh/eo-vae}{https://github.com/nilsleh/eo-vae}}

\section{Methodology}
\subsection{Dataset}
We choose the TerraMesh dataset \citep{blumenstiel2025terramesh} to train and evaluate our EO-VAE because this allows for a fair comparison against the TerraMind tokenizers which have been trained on the same data. We follow the same z-score normalization scheme of TerraMind but also find that there is a misalignment in the Sentinel 2 data based on a new processing mode introduced in January 2022, as well as missing data for some modalities. More information is provided in the appendix. We do experiments with both a native and a corrected data corpus for which results are provided in the appendix. We train and evaluate our model on the native Sentinel-2 L2A, and Sentinel-1 RTC modalities for which data is complete and pretrained tokenizers available. Due to storage demands we are only able to train on a subset of the TerraMesh data (first 25 shards). We generate a separate test split from the complete TerraMesh validation splits, where shards 0-6 are validation and shards 6-8 the test split and use an image size of 256x256px.

\subsection{Model}
\begin{figure}
    \centering
    \includegraphics[width=0.7\linewidth]{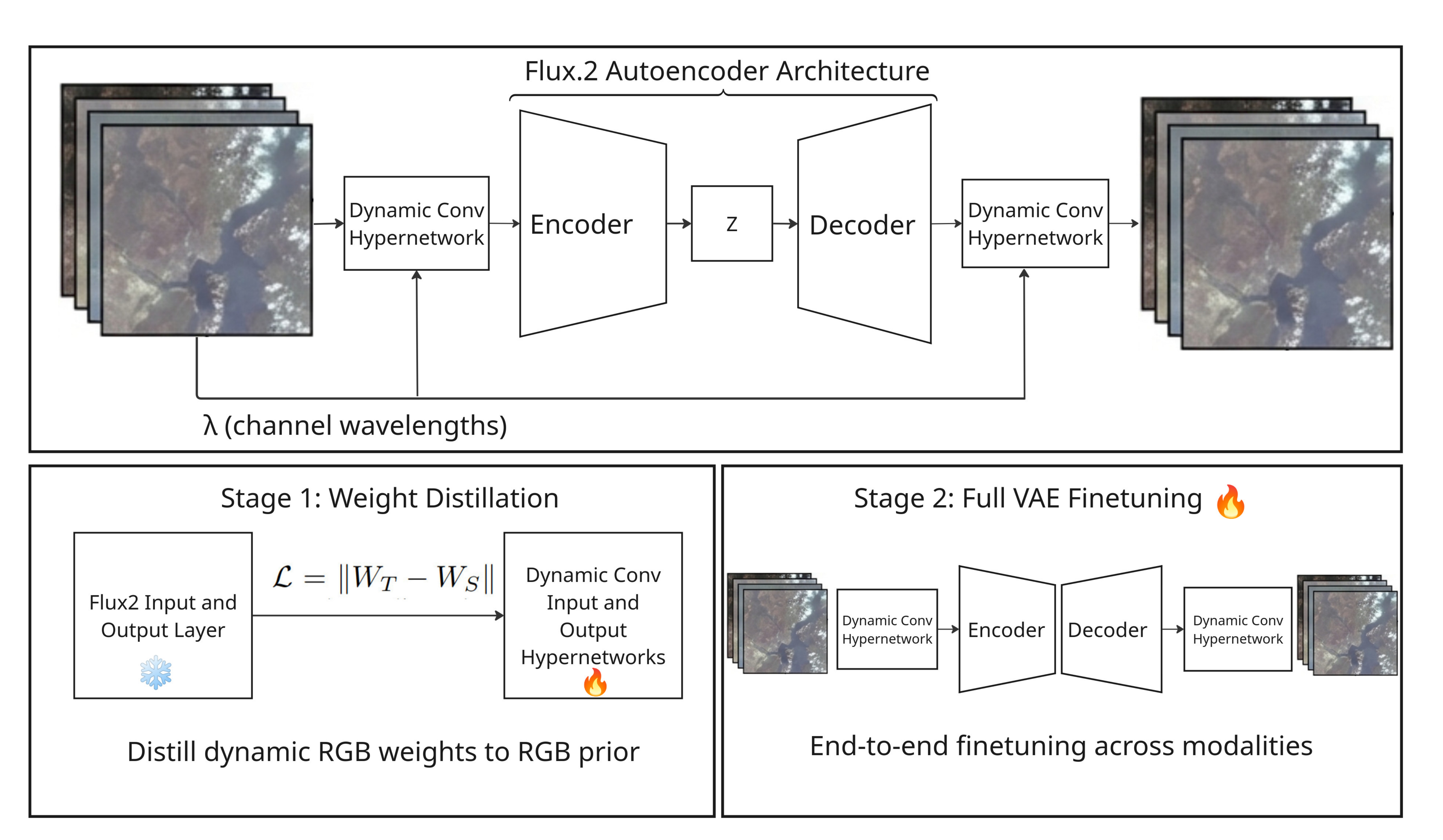}
    \caption{EO-VAE Architecture and Training Regime. The first and last convolutional layer of the Flux.2 Autoencoder architecture are replaced with dynamic convolution hypernetworks \citep{xiong2024neural}. After weight distillation of the frozen Flux.2 convolutional weights, we finetune end-to-end on the multimodal TerraMesh dataset.}
    \label{fig:eo-vae}
\end{figure}

We use the recently introduced Flux.2 Autoencoder \citep{flux2} as a base architecture and pretrained checkpoint. To accommodate a flexible number of input channels from various satellite modalities, we replace the first and last convolutional layer with dynamic hypernetworks convolutional layers that generate the convolutional weights conditioned on the channel wavelengths as proposed in the DOFA model \citep{xiong2024neural}. This base model is able to reconstruct flexible number of channel combinations. Overall our training regime consists of two stages, which is depicted in Figure \ref{fig:eo-vae}:

\begin{enumerate}
    \item First, we use weight distillation \citep{lin2021weight} of the Flux.2 autoencoder's first and last convolutional layer (teacher) into the dynamic weight layers (student), by minimizing $\mathcal{L}=\left\|W_T-W_S \right\|$ where $W_T$ are the teacher weights and $W_S$ are the student weights through gradient descent optimization. We find this distillation to be crucial for faster convergence. The RGB channel provides a strong prior before exposing them to multispectral data.
    \item Second, we conduct full finetuning across all three modalities via pixel-wise reconstruction loss.
\end{enumerate}

For the reconstruction objective, let $x \in \mathbb{R}^{C \times H \times W}$ denote an input satellite image with $C$ spectral channels and corresponding channel wavelengths $\lambda = (\lambda_1,\dots,\lambda_C)$. 
Let $E_{\theta_E}(\cdot;\lambda)$ and $D_{\theta_D}(\cdot;\lambda)$ denote the encoder and decoder of the Flux.2 autoencoder, where the first and last convolutional layers are replaced by dynamic hypernetwork layers that generate convolutional weights conditioned on $\lambda$. The reconstructed output is given by
\begin{equation}
\hat{x} = D_{\theta_D}\!\left(E_{\theta_E}(x;\lambda);\lambda\right)
\end{equation}

The objective is to minimize a reconstruction loss over the training dataset $\mathcal{D}$:
\begin{equation}
\mathcal{L}_{\mathrm{rec}}(\theta_E,\theta_D)
=
\mathbb{E}_{(x,\lambda)\sim\mathcal{D}}
\left[
\ell(x,\hat{x})
\right],
\end{equation}
where $\ell(\cdot,\cdot)$ is a per-pixel reconstruction loss. 
We use an equally weighted Charbonier \citep{charbonnier1994two} and multiscale structure similarity index \citep{wang2003multiscale} loss. All experiments were run on a 48GB NVIDIA RTX A6000. 

\section{Results}
We evaluate EO-VAE on two capabilities: (1) high-fidelity reconstruction of multi-modal satellite imagery, and (2) its utility as a frozen latent tokenizer for downstream generative tasks. To capture complementary components of image prediction quality, we use RMSE, PSNR, SSIM, and SAM. Additionally, for the S2 modality, we also evaluate the reconstructed Normalized Difference Vegetation Index (NDVI) in terms of Mean Absolute Error (MAE) to assess physical consistency.

\subsection{Reconstruction}

\begin{figure}[t]
    \centering
    \begin{subfigure}[b]{0.48\linewidth}
        \centering
        \includegraphics[width=\linewidth]{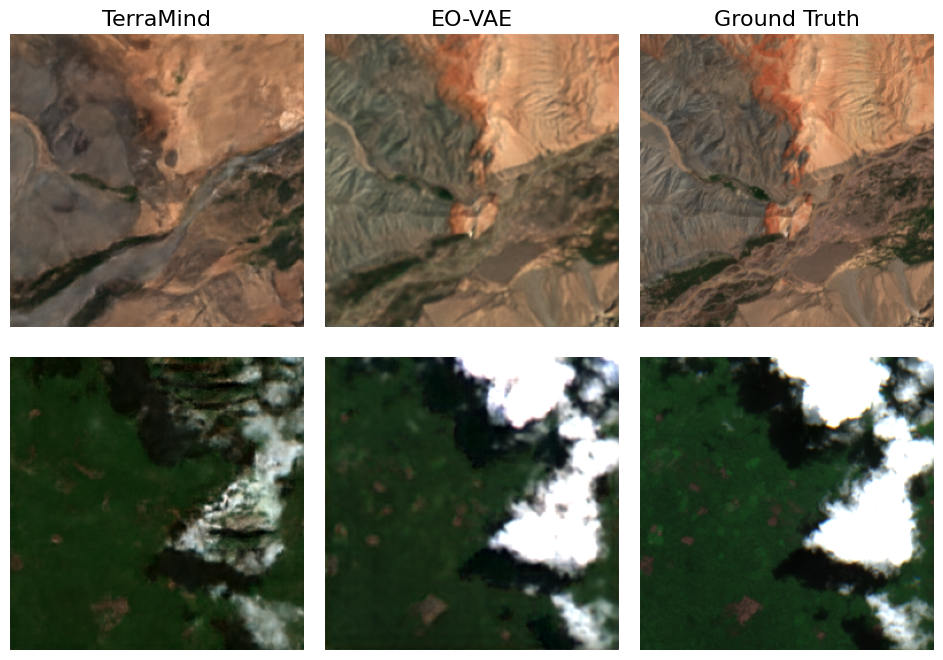}
        \caption{S2L2A modality reconstruction.}
        \label{fig:qual-sl2a-left}
    \end{subfigure}
    \hfill
    \begin{subfigure}[b]{0.48\linewidth}
        \centering
        \includegraphics[width=\linewidth]{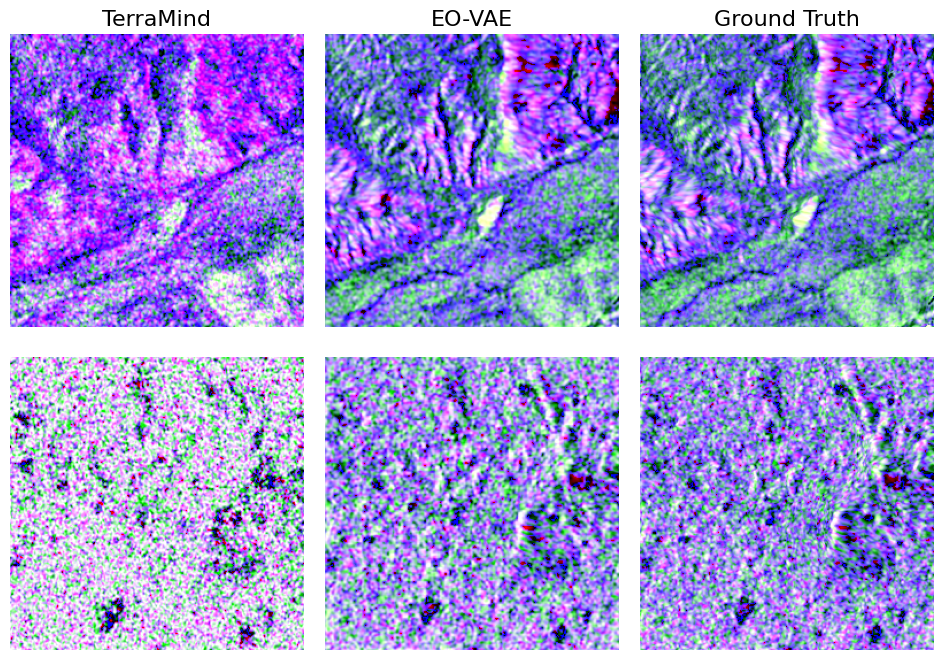}
        \caption{S1RTC modality reconstruction.}
        \label{fig:qual-sl2a-right}
    \end{subfigure}
    \caption{Qualitative samples of reconstructed modalities. EO-VAE reconstructs details in both modalities better than the TerraMind tokenizers.}
    \label{fig:qual-sl2a}
\end{figure}

Table \ref{tab:reconstruction_performance} demonstrates that EO-VAE substantially outperforms the TerraMind Tokenizers across all metrics for both the S2L2A and S1RTC modality. On the S2L2A modality, EO-VAE achieves a PSNR of 42.80 dB, nearly 20 dB higher than TerraMind (22.95 dB). This quantitative gap aligns with the qualitative results in Figure \ref{fig:qual-sl2a}, where EO-VAE preserves high-frequency details significantly better for both modalities. Additionally, our model achieves a substantial 3.5$\times$ reduction in NDVI MAE of reconstructed S2L2A images, and therefore better captures this crucial inter-band ratio.

\begin{table*}[h]
\centering
\resizebox{\linewidth}{!}{
\begin{tabular}{lccccccccc}
\toprule
\textbf{Model} & \multicolumn{4}{c}{\textbf{S1RTC}} & \multicolumn{5}{c}{\textbf{S2L2A}} \\
\cmidrule(lr){2-5} \cmidrule(lr){6-10}
 & RMSE$\downarrow$ & PSNR$\uparrow$ & SSIM$\uparrow$ & SAM$\downarrow$ & RMSE$\downarrow$ & PSNR$\uparrow$ & SSIM$\uparrow$ & SAM$\downarrow$ & NDVI-MAE$\downarrow$ \\
\midrule
EO-VAE & \textbf{0.1401} & \textbf{37.23} & \textbf{0.9372} & \textbf{0.1601} & \textbf{0.0686} & \textbf{42.80} & \textbf{0.9720} & \textbf{0.0842} & \textbf{0.0410} \\
TerraMind & 0.6711 & 23.65 & 0.2803 & 0.7285 & 0.7004 & 22.95 & 0.7543 & 0.3568 & 0.1403 \\
\bottomrule
\end{tabular}}
\caption{Reconstruction performance across modalities. EO-VAE outperforms the TerraMind tokenizers across both modalities and all metrics.}
\label{tab:reconstruction_performance}
\end{table*}

\subsection{Downstream Task: Latent Super Resolution}
To demonstrate the utility of our compressed latent space, we evaluate EO-VAE as a fixed tokenizer for a Latent Diffusion Model (LDM) super-resolution task on the Cross-Sensor Sen2NAIP dataset \citep{aybar2024sen2naip}. This dataset consists of spatially aligned Sentinel-2 and NAIP imagery with RGBN bands and a resolution factor of 4 from 128 to 512 pixels. We adopt a checkerboard-style geospatial split inspired by MOSAIK \citep{rolf2021generalizable} to ensure spatial separation between training, validation, and test regions, resulting in 2417/288/146 data points per split. 

We formulate the task as a Latent Diffusion Model (LDM). Both low- and high-resolution images are encoded into the latent space using a frozen autoencoder. We train a standard UNet backbone \citep{ronneberger2015u} to predict the high-resolution latents, conditioned on the upsampled low-resolution latents via concatenation. We use the EDM \citep{karras2022elucidating} diffusion model implemented in the \texttt{azula} library.\footnote{See \href{https://azula.readthedocs.io/stable/}{https://azula.readthedocs.io/stable/}.} The EDM loss is defined as $\frac{\alpha_t^2 + \sigma_t^2}{\sigma_t^2} || \mu_\phi(x_t) - x ||^2$, where $\alpha$ and $\sigma$ denote the noise parameters from the variance preserving schedule used by \cite{songscore}. For inference we use DDIM sampler with 50 steps \citep{songdenoising}.

\begin{figure}
    \centering
    \includegraphics[width=0.9\linewidth]{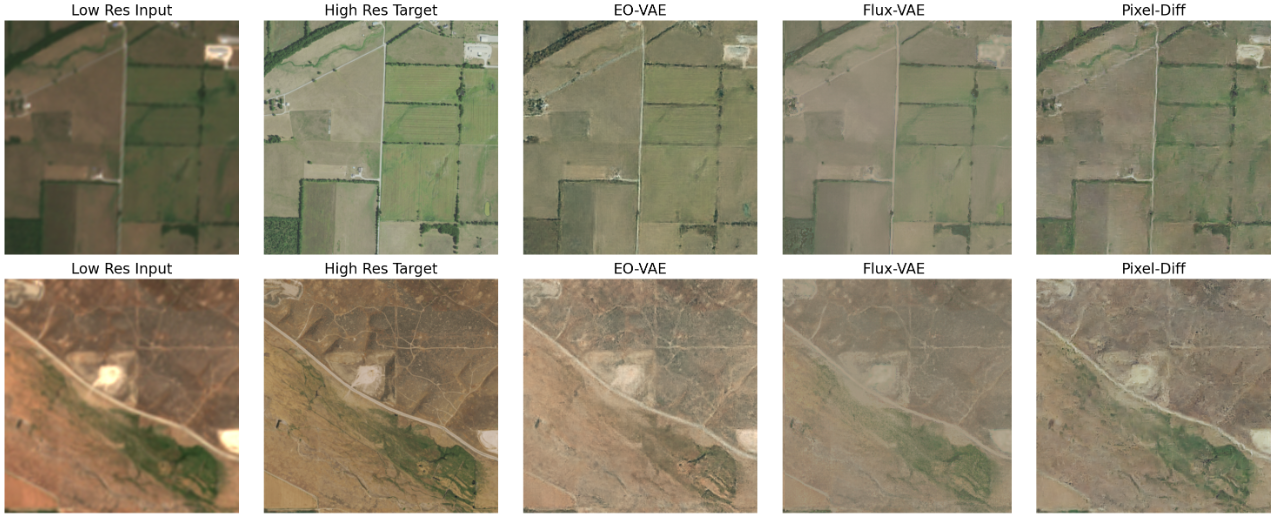}
    \caption{Qualitative Results between EO-VAE and Flux-VAE for reconstructed super-resolution predictions.}
    \label{fig:super-res-comparison}
\end{figure}

This experiment highlights a practical limitation of existing foundation tokenizers. TerraMind cannot support this task, as no pretrained model is available for the RGBN modality, and training one from scratch would be computationally prohibitive. EO-VAE, in contrast, naturally adapts to 4-channel input without architectural modification. We compare EO-VAE against two available baselines: (i) a pretrained Flux.2 VAE restricted to RGB channels (dropping NIR), since the original model cannot process multispectral inputs, and (ii) a native pixel-space diffusion model trained directly on low- and high-resolution images.

Quantitative results are reported in Table~\ref{tab:sr_results_with_compute}. EO-VAE achieves performance on par with the frozen RGB Flux.2 VAE, indicating that extending the tokenizer to multispectral input does not substantially degrade generative fidelity. At the same time, both latent-based approaches clearly outperform the pixel-space diffusion baseline, which is also reflected in the qualitative comparison in Figure~\ref{fig:super-res-comparison}. Beyond reconstruction quality, Table~\ref{tab:sr_results_with_compute} demonstrates the computational advantage of latent diffusion, with substantial efficiency gains compared to operating directly in pixel space.


\begin{table}[h]
\centering
\resizebox{\linewidth}{!}{
\begin{tabular}{llccccrccr}
    \toprule
    Model 
    & Bands
    & PSNR$\uparrow$ 
    & SSIM$\uparrow$ 
    & RMSE$\downarrow$ 
    & SAM$\downarrow$ 
    & \makecell{Time \\ (ms)$\downarrow$} 
    & \makecell{Throughput \\ (img/s)$\uparrow$} 
    & \makecell{Peak Memory \\ (GB)$\downarrow$} 
    & \makecell{Params (M) \\ Total (Diffusion)} \\
    \midrule
    EO-VAE 
    & RGB+NIR
    & 21.60 & 0.6234 & 0.0836 & 0.0672 
    & 389.7 & 2.57 & 1.53 & 106.5 (11.0) \\


    Flux.2 VAE 
    & RGB
    & \textbf{21.94} & \textbf{0.6434} & \textbf{0.0810} & \textbf{0.0609} 
    & \textbf{374.7} & \textbf{2.67} & \textbf{1.41} & 95.0 (11.0) \\

    PIXELDiff 
    & RGB+NIR
    & 21.76 & 0.3437 & 0.7616 & 0.6905 
    & 7097.9 & 0.14 & 1.69 & 10.8 (10.8) \\
    \bottomrule
\end{tabular}}
\caption{Test set metrics on Super-Resolution experiments. EO-VAE performs on par with RGB Flux.2 model. Computational inference metrics are averaged over 50 iterations. Latent diffusion approaches are ~18x more efficient measured by Time (ms) than pixel space approach.}
\label{tab:sr_results_with_compute}
\end{table}


\vspace{-1em}
\section{Conclusion and Future Work}

We presented EO-VAE, a modality-agnostic tokenizer designed for multispectral Earth Observation data. For reconstruction, we compared against TerraMind foundation tokenizers and observed consistent improvements across all reported modalities and metrics, including improved preservation of NDVI error, despite training on a substantially smaller dataset. These results demonstrate that EO-VAE maintains high-fidelity reconstruction while better capturing spectral structure. For downstream generative modeling, we evaluated EO-VAE in a latent diffusion super-resolution task and compared against both pixel-space diffusion and a strong RGB-only pretrained model. While large-scale RGB models like Flux.2 remain highly competitive on RGB-only benchmarks, they are not directly applicable to heterogeneous multisensor data without retraining. EO-VAE achieves competitive performance while natively supporting multispectral inputs and enabling unified latent modeling across channel compositions. In addition, operating in the EO-VAE latent space provides an approximately 18$\times$ inference speedup compared to pixel-space diffusion. Overall, EO-VAE bridges high-fidelity reconstruction and modality flexibility, offering a pathway for a practical tokenizer for multisensor Earth Observation pipelines. Future work includes scaling across additional sensors and resolutions, improving perceptual quality, and extending the framework to spatio-temporal 3D architectures for time-series modeling.

\bibliography{iclr2026_conference}
\bibliographystyle{iclr2026_conference}

\newpage
\appendix

\section{Appendix}
\subsection{TerraMesh S1-GRD Modality}

At the point of submission, there was no matching validation data for the S1-GRD modality available \href{https://huggingface.co/datasets/ibm-esa-geospatial/TerraMesh/tree/main/val/S1GRD}{HuggingFace repo}.
\subsection{TerraMesh Sentinel 2 S2L2A modality}
When inspecting the Sentinel-2 L2A surface reflectance imagery in the TerraMesh dataset, we found a clear inconsistency that we found was not clearly documented. On January 25, 2022, ESA introduced a new processing baseline update (Baseline 04.00), which changes the way Sentinel-2 pixel values are represented that shifted the data range by adding a constant offset to allow negative reflectance values to be encoded. In practice this means that for imagery processed under this new baseline, the digital numbers (DNs) include values down to about –1000, whereas earlier imagery sticks to values near zero for very low reflectance and NoData. \cite{blumenstiel2025terramesh} state that “the +1000 offset is removed from post-2022 data” and reports a value range of [0, 10000] for S2L2A. However, based on data analysis, it doesn’t appear to have been applied consistently: histograms and time series of minimum pixel values show two distinct populations of Sentinel-2 values, with post-baseline images exhibiting significant density at negative values that line up with the offset.

\begin{figure}[htpb]
    \centering
    \includegraphics[width=0.8\linewidth]{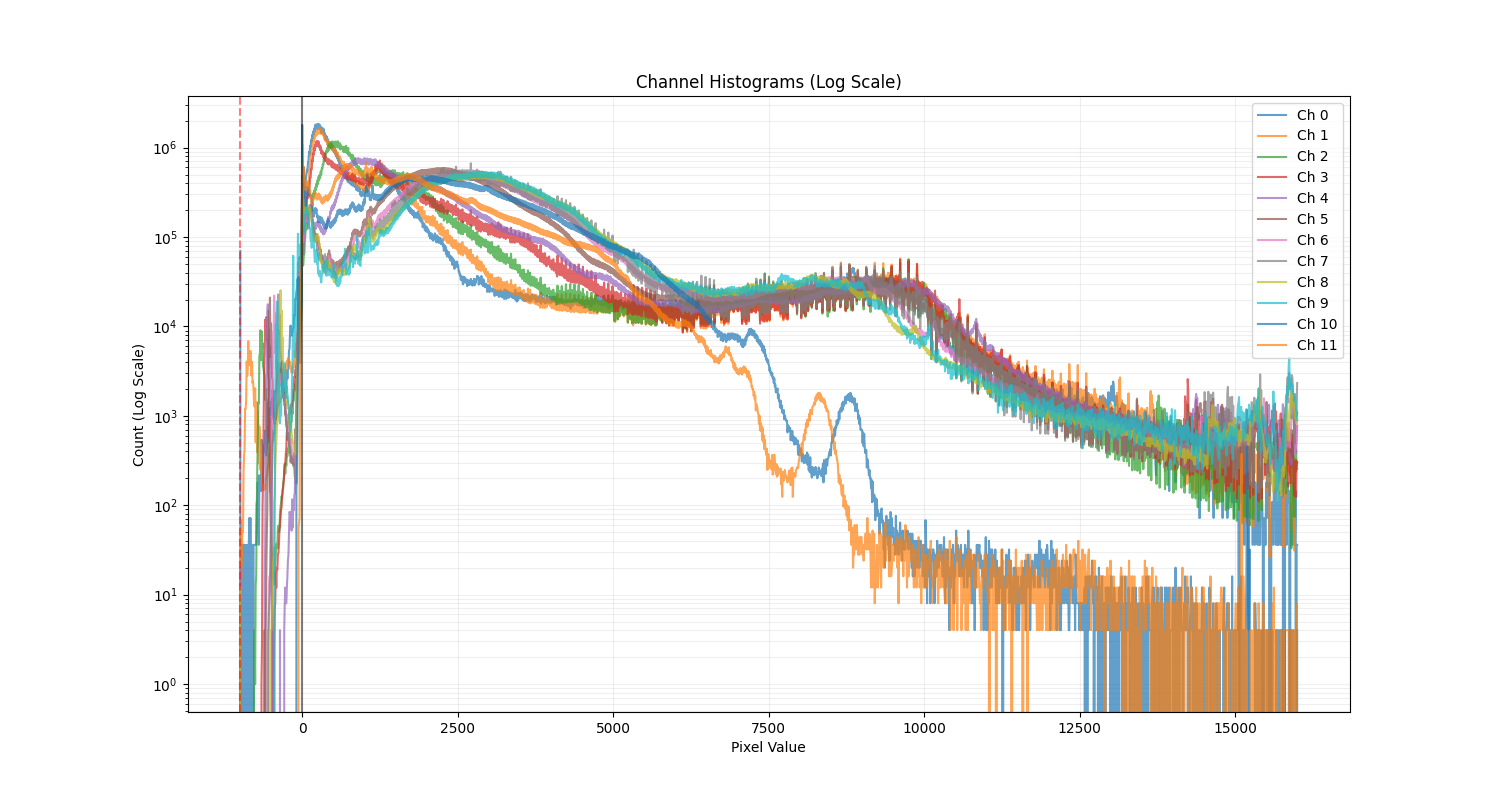}
    \caption{channelwise histogram of raw unnormalized data for the S2L2A modality, showing the range of \textgreater 10000.}
    \label{fig:s2l2a-data-time}
\end{figure}

\begin{figure}[htpb]
    \centering
    \includegraphics[width=0.8\linewidth]{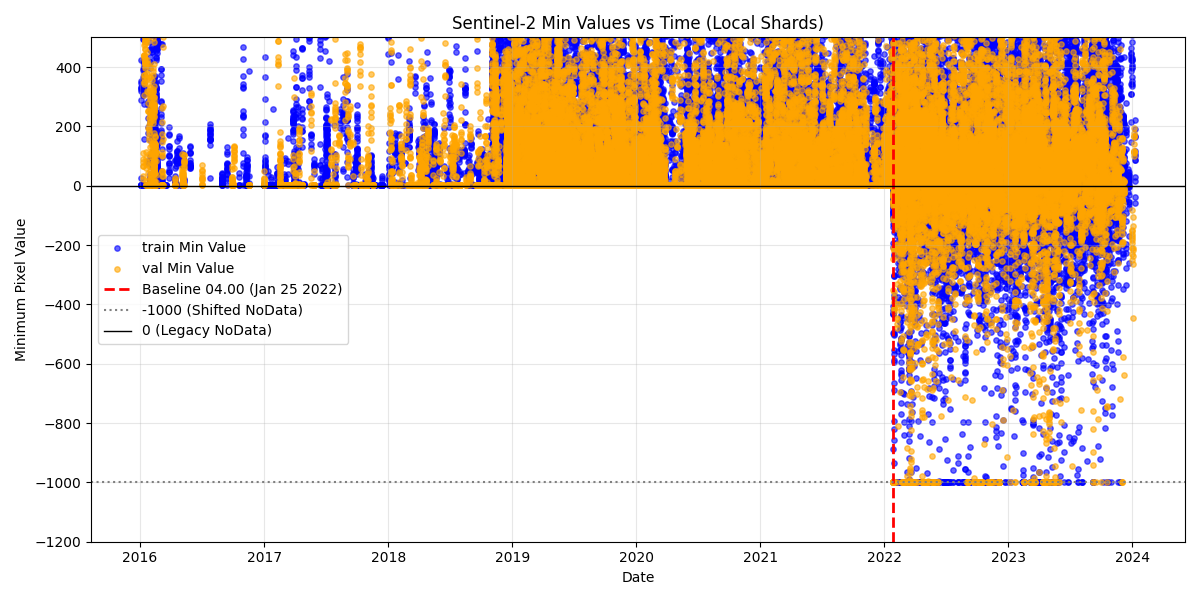}
    \caption{Minimum sample values plotted across time. The processing baseline change on January 22, 2022 becomes clearly visible.}
    \label{fig:s2l2a-data-time}
\end{figure}

This leads to systematic differences in the basic statistics of the image data across the pre- and post-baseline change periods. The discontinuity is visible in the plots of minimum values over time as a sharp transition around the baseline change date, and the channel histograms highlight that negative values (around –1000) are concentrated in the later imagery. Including both conventions in the same dataset without adjustment therefore introduces a data inconsistency in the Sentinel-2 pixel value distributions. However, we find that correcting this bias does not change the results in a statistically significant manner. 

\subsection{Results with corrected S2L2A data}

The following section lists results with the corrected data, where we have harmonized the data and computed new normalization statistics. Interestingly enough, the reconstruction results do not seem to improve with a EO-VAE trained on this corrected data.

\begin{table*}[htbp]
\centering
\small
\begin{tabular*}{\textwidth}{@{\extracolsep{\fill}}lcccccccc}
\toprule
\textbf{Model} & \multicolumn{4}{c}{\textbf{S1RTC}} & \multicolumn{4}{c}{\textbf{S2L2A}} \\
\cmidrule(lr){2-5} \cmidrule(lr){6-9}
 & RMSE$\downarrow$ & PSNR$\uparrow$ & SSIM$\uparrow$ & SAM$\downarrow$ & RMSE$\downarrow$ & PSNR$\uparrow$ & SSIM$\uparrow$ & SAM$\downarrow$ \\
\midrule
EO-VAE & \textbf{0.1401} & \textbf{37.23} & \textbf{0.9372} & \textbf{0.1601} & \textbf{0.0686} & \textbf{42.80} & \textbf{0.9720} & \textbf{0.0842} \\
EO-VAE* & 0.1779 & 35.22 & 0.9010 & 0.1940 & 0.0904 & 37.58 & 0.9383 & 0.1135 \\
TerraMind & 0.6711 & 23.65 & 0.2803 & 0.7285 & 0.7004 & 22.95 & 0.7543 & 0.3568 \\
\bottomrule
\end{tabular*}
\caption{Reconstruction Performance Across Modalities. EO-VAE* denotes the version trained with the corrected aligned S2L2A data corpus.}
\label{tab:recon_with_correction}
\end{table*}

In contrast, on the downstream task for super-resolution we observe, that the performance is slightly higher. In Table \ref{tab:sr_results_with_compute_2} shows

\begin{table}[h]
\centering
\resizebox{\linewidth}{!}{
\begin{tabular}{llccccrrrr}
    \toprule
    Model 
    & Bands
    & PSNR$\uparrow$ 
    & SSIM$\uparrow$ 
    & RMSE$\downarrow$ 
    & SAM$\downarrow$ 
    & \makecell{Time \\ (ms)$\downarrow$} 
    & \makecell{Throughput \\ (img/s)$\uparrow$} 
    & \makecell{Memory \\ (GB)$\downarrow$} 
    & \makecell{Params \\ (M)} \\
    \midrule
    EO-VAE 
    & RGB+NIR
    & 21.60 & 0.6234 & 0.0836 & 0.0672 
    & 389.7 & 2.57 & 1.53 & 106.5 (11.0) \\

    EO-VAE* 
    & RGB+NIR
    & \textbf{22.17} & \textbf{0.6556} & \textbf{0.0785} & \textbf{0.0584} 
    & 389.7 & 2.57 & 1.53 & 106.5 (11.0) \\

    Flux.2 VAE 
    & RGB
    & 21.94 & 0.6434 & 0.0810 & 0.0609 
    & 374.7 & 2.67 & 1.41 & 95.0 (11.0) \\

    PIXELDiff 
    & RGB+NIR
    & 21.76 & 0.3437 & 0.7616 & 0.6905 
    & 7097.9 & 0.14 & 1.69 & 10.8 (10.8) \\
    \bottomrule
\end{tabular}}
\caption{Test set metrics on Super-Resolution experiments. EO-VAE* denotes version trained on corrected S2L2A data corpus. Computational metrics are averaged over 50 iterations.}
\label{tab:sr_results_with_compute_2}
\end{table}

\begin{figure}
    \centering
    \includegraphics[width=0.9\linewidth]{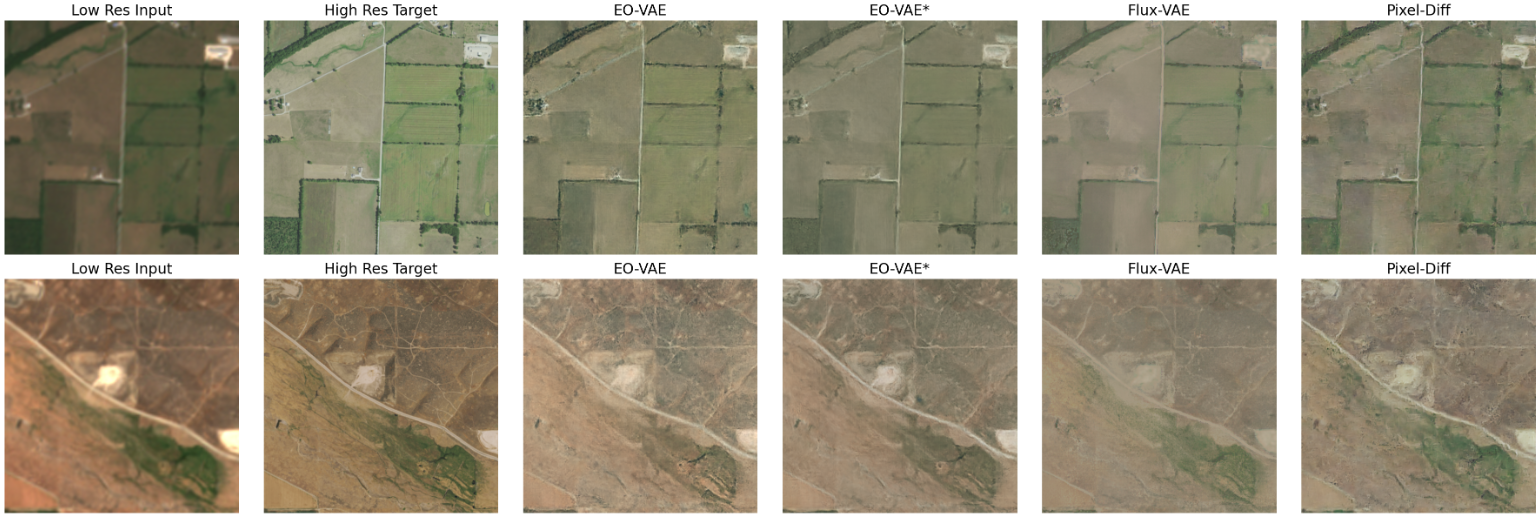}
    \caption{Qualitative Results between EO-VAE and Flux-VAE for reconstructed super-resolution predictions}
    \label{fig:super-res-comparison-2}
\end{figure}

Fig \ref{fig:super-res-comparison-2} shows a qualitative example with very similar visual results.

\end{document}